\documentclass[10pt,twocolumn,letterpaper]{article}

\usepackage{cvpr}
\usepackage{times}
\usepackage{epsfig}
\usepackage{graphicx}
\usepackage{amsmath}
\usepackage{amssymb}
\usepackage{multirow}
\usepackage{float}
\usepackage{times}
\usepackage{tabularx}
\usepackage{paralist}
\usepackage{cite}
\usepackage[nopar]{lipsum}
\usepackage{diagbox} 
\usepackage{cancel}
\usepackage{subfigure}

\usepackage{paralist}

\usepackage[pagebackref=true,breaklinks=true,letterpaper=true,colorlinks,bookmarks=false]{hyperref}

\usepackage{xspace}

\newcommand{\mbf}{\mathbf}

\newcommand{\img}{\ensuremath{\mbf I}\xspace}
\newcommand{\deltaGaze}{\ensuremath{\Delta \mbf g}\xspace}
\newcommand{\headPose}{\ensuremath{\mbf h}\xspace}

\newcommand{\warpX}{\ensuremath{\mbf m_{x}}\xspace}
\newcommand{\warpY}{\ensuremath{\mbf m_{y}}\xspace}
\newcommand{\warpXY}{\ensuremath{\mbf m_{x,y}}\xspace}

\newcommand{\RNetParameter}{\ensuremath{\mbf \theta}\xspace}
\newcommand{\RNet}{\ensuremath{\mbf R}\xspace}
\newcommand{\gazeNet}{\ensuremath{\mbf E_{\phi}}\xspace}

\newcommand{\gridSampler}{\ensuremath{s}\xspace}

\newcommand{\gtI}{\ensuremath{\mbf G_{\img}}\xspace}

\newcommand{\segI}{\ensuremath{\mbf S_{\img}}\xspace}

\newcommand{\gtSeg}{\ensuremath{\mbf S_{\mbf G_{\img}}}\xspace}

\newcommand{\gazeLabel}{\ensuremath{\mbf g}\xspace}

\newcommand{\numSample}{\ensuremath{n}\xspace}

\newcommand{\redTime}{\ensuremath{t}\xspace}

\newcommand{\difNet}{\textit{DiffNet}\xspace}
\newcommand{\vgg}{\textit{vgg16}\xspace}

\newcommand{\LinAdpt}{\textit{LinAdap}\xspace}
\newcommand{\SVRAdpt}{\textit{SVRAdap}\xspace}
\newcommand{\Finetune}{\textit{FTAdap}\xspace}
\newcommand{\AugFinetune}{\textit{RedFTAdap}\xspace}
\newcommand{\AugFinetuneNoDA}{\textit{RedFTAdap-noDA}\xspace}

\newcommand{\FT}{\textit{FT}\xspace}
\newcommand{\Red}{\textit{Red}\xspace}

\newcommand{\mypartitle}[1]{\vspace*{0.5mm}{\noindent {\bf #1}}}

\def\cvprPaperID{6510} % *** Enter the CVPR Paper ID here

\cvprfinalcopy % *** Uncomment this line for the final submission
\begin{document}

%%%%%%%%% TITLE
% \title{Improving User-Specific Gaze Estimation via Gaze Redirection Synthesis}
\title{\vspace*{-25mm}
  {\footnotesize \rm  Submitted to CVPR 2019 on Nov 15th, 2018. Accepted to CVPR 2019.}~\\[10mm]
Improving Few-Shot User-Specific Gaze Adaptation \\ via Gaze Redirection Synthesis\\[-2mm]}

% \author{First Author\\
% Institution1\\
% Institution1 address\\
% {\tt\small firstauthor@i1.org}
% % For a paper whose authors are all at the same institution,
% % omit the following lines up until the closing ``}''.
% % Additional authors and addresses can be added with ``\and'',
% % just like the second author.
% % To save space, use either the email address or home page, not both
% \and
% Second Author\\
% Institution2\\
% First line of institution2 address\\
% {\tt\small secondauthor@i2.org}
% }

\author{
  Yu Yu, \  \ Gang Liu,  \ \ Jean-Marc Odobez\\[-0.25mm]
Idiap Research Institute, CH-1920, Martigny, Switzerland\\[-0.25mm]
EPFL, CH-1015, Lausanne, Switzerland \\ [-0.25mm]
 \{yyu, gang.liu, odobez\}@idiap.ch\\[-2mm]
}

\maketitle
%\thispagestyle{empty}
%%%%%%%%% ABSTRACT

\begin{abstract}
\vspace*{-2mm}
   As an indicator of human attention
  gaze is a subtle behavioral cue which can be exploited in many applications.
  However, inferring 3D gaze direction is challenging even for deep neural networks given
  the lack of large amount of data  (groundtruthing  gaze is expensive
  and existing datasets use different setups) and the inherent
  presence of gaze biases due to person-specific difference.
  In this work, we address the problem of person-specific gaze model adaptation
  from only a few reference training samples. 
  The main and novel idea is to improve gaze adaptation by generating additional training samples through
  the synthesis of gaze-redirected eye images from  existing reference samples. 
  In doing so, our contributions are threefold:
  (i) we design our gaze redirection framework from synthetic data,
  allowing us to benefit from aligned training sample pairs to predict accurate
  inverse mapping fields;
  (ii) we proposed a self-supervised approach for domain adaptation;
  (iii) we exploit the gaze redirection to improve the performance of person-specific gaze estimation.
%  for the first time to our best knowledge, 
%
  Extensive experiments on two public datasets demonstrate the validity
  of our gaze retargeting and gaze estimation framework.
%
% \keywords{Gaze Estimation \and Gaze Redirection \and GAN}
\end{abstract}

% - Gaze important human behaviour
% - Problem of gaze, expensive to annotate, personal bias system bias between databases
% - Methods proposed for domain adptation, domain difference, label bias
% - Gaze redirection from few samples

\vspace*{-2mm}

\section{Introduction}

\vspace*{-1mm}

Gaze, as a subtle non-verbal human behaviour, not only
indicates the visual content people perceive but also conveys information about
the level of attention, mental state or even higher level psychological constructs of human.
As a consequence,
%means of information exchange between human and the world, the
gaze cues have  been exploited in many areas like social interaction analysis~\cite{Ishii:2016:PNS:2896319.2757284},
stress analysis~\cite{huang2016stressclick}, human robot interaction (HRI)~\cite{Andrist:2014:CGA:2559636.2559666,Moon:2014:MMI:2559636.2559656}, the emerging Virtual Reality  industry~\cite{patney2016towards,padmanaban2017optimizing},
and they are expected to find a wide range of application in mobile interactions
with smart phones~\cite{Krafka2016,tonsen17_imwut,huang2015tabletgaze}.

\begin{figure}[tb]
  \centering
  \vspace*{-0.5em}\includegraphics[height=35mm]{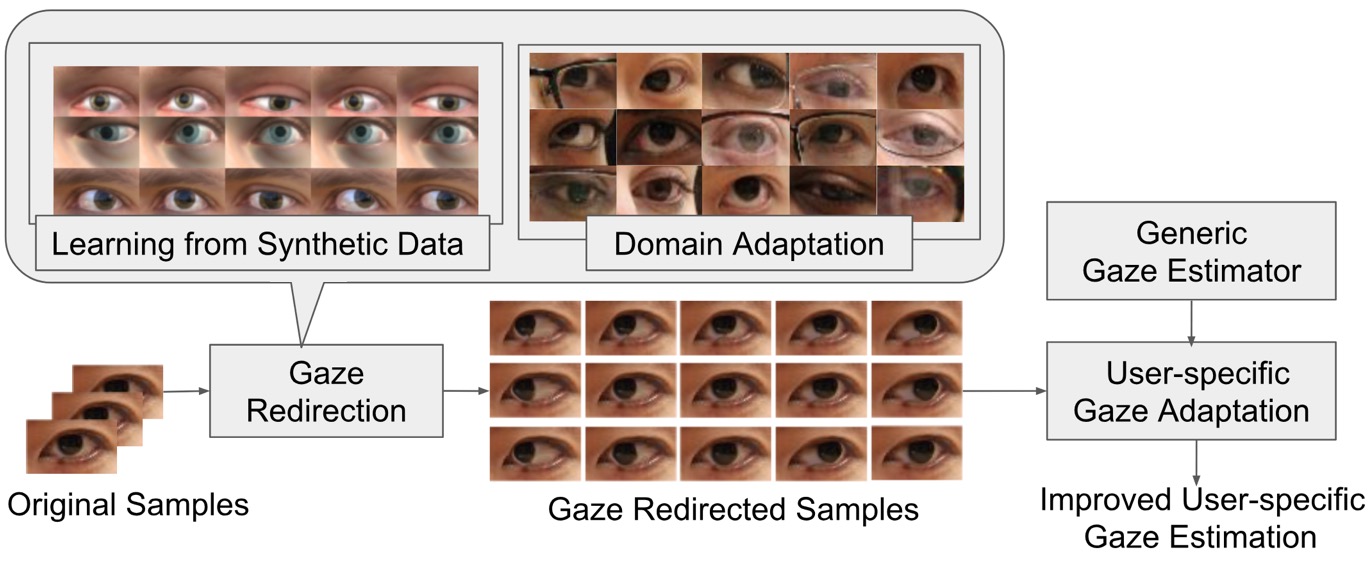}
  \vspace*{-2mm}
  \caption{Approach overview.
    A few reference eye images (with gaze ground truth) from a user are used as input to a gaze redirection synthesis
    module to generate further training samples. The latter (and reference samples) are used to fine-tune
    a generic gaze estimator to obtain a user-specific gaze estimator.
  }
  \label{fig:main_idea}
\vspace*{-3mm}
\end{figure}

However, gaze extraction from non invasive visual sensors is challenging
and has attracted an increased amount of research in recent years.
Approaches can be classified in two general categories: geometric based methods (GBM) and appearance based methods (ABM).
The former ones rely on a geometrical model of eyes whose parameters can be inferred
from localized eye landmarks like iris or eye corners.
Although they can be very accurate, they usually require high resolution data to reliably extract eye features.
The latter ABM  ones directly learn a mapping from the eye images to the gaze directions
and have been shown to be more robust against low eye image  resolution
and other variability factors (illumination, head pose, gaze range,...).
Nevertheless, in spite of recent progresses partly due to the use of deep neural networks~\cite{zhang2015appearance,Zhang2017a,Krafka2016,Fischer2018,Liu2018,Park2018a}, 
vision based gaze estimation is still a challenging and open problem due to at least three main factors:
\begin{compactitem}
\item \textbf{Lack of data.}
  The sizes of benchmark gaze datasets~\cite{funes2014eyediap,zhang2015appearance,smith2013gaze} are relatively small compared
  to other vision tasks like image classification, since accurate gaze annotation is complex and expensive.
  To address the lack of data,  domain adaptation methods~\cite{shrivastava2017learning}
  have proposed to use  synthetic images for training,
  but completely eliminating the domain discrepancies between real and synthetic eye images is hard.
 % On the other hand, the existing gaze datasets differ in elements like input (single eye images vs full face images) and output (3D gaze directions vs 2D gaze fixation points), making it hard to merge various datasets for training. 
% Since the annotation of gaze is complex and expensive, the sizes of the benchmark gaze datasets~\cite{funes2014eyediap,zhang2015appearance,smith2013gaze} are relatively small compared with other vision tasks like image classification, object detection or even body landmark localization. Furthermore, the settings and resolutions of these databases differ a lot (e.g. 3D gaze direction VS 2D gaze fixation, low resolution VS high resolution), making it hard to merge them for training. Although some domain adaptation methods~\cite{shrivastava2017learning} have been proposed to refine synthetic images for training, they can not eliminate the domain difference completely.

\item \textbf{Systematic bias.}
  Existing gaze datasets usually use different gaze coordinate systems and data pre-processing methods,
  in particular for geometric normalization (rectification) relying on different head pose estimators.
  This introduces a between-dataset systematic bias regarding the gaze groundtruth~\cite{yudeep}.

\item \textbf{Person-specific bias.} Liu et al.~\cite{Liu2018} legitimaly argue that gaze can not be
  fully estimated from the visual appearance since the alignment difference between the optical axis (the line connecting the eyeball center and the pupil center) 
  and the visual axis (the line connecting the fovea and the nodal point~\cite{FunesMora2014}) is person specific, and vary within -2 to 2 degrees across the population.
  Therefore, it is not optimal to train a single generic model for accurate cross-person gaze estimation. 
\end{compactitem}

\vspace*{1mm}

In this paper, we focus on the problem of person-specific gaze adaptation which has not received
enough attention compared to cross person gaze estimation.
More specifically, the aim is to only rely on few samples since collecting  training samples for a new subject is expensive.
In this context, a first and interesting result that we show is that a direct and simple fine tuning of a neural network gaze regressor
can improve person-specific gaze estimation by a good margin, even if the number of person-specific samples is as small as 9.
We then propose to further improve the performance of such gaze adaptation method by
using as additional training data gaze-redirected samples synthesized from the
given reference samples, as illustrated in Fig.~\ref{fig:main_idea}.
%
%Our idea is inspired by the works~\cite{Wood2018,Kononenko2017,Ganin2016,Hsu2018}
%of Gaze Redirection which is originally for gaze correction in video conference. 
%
Compared with domain adaptation methods like SimGAN~\cite{shrivastava2017learning},
which work by retargeting synthetic images into subject specific eye images,
we firmly believe that a gaze redirection framework relying on reference eye images and user defined gaze changes (redirection angles)
can generate samples with more realistic appearance (since they are directly derived from real eye images of the subject)
and more reliable groundtruth (less systematic and person-specific bias),
thus demonstrating better performance when used for person-specific gaze adaptation.
By investigating the aboves ideas, we make the following contributions:
%
% Compared with images generated by domain adaptation methods like
% SimGAN~\cite{shrivastava2017learning} which refines synthetic samples, 
% we firmly believe that a gaze-redirected sample (whose label is the addition of the reference image groundtruth and the user defined gaze change) which is synthesized from a reference image and a user defined gaze change (redirection angle) can be more realistic and suffers from less systematic bias and personal bias (since the groundtruth of the new samples is based on a reference label), thus demonstrating better performance when used for user-specific gaze adaptation. By investigating the aboves ideas, we make the following contributions:
%
\begin{compactitem}
%  \item \textbf{Synthesis learned gaze redirection.} 
  \item \textbf{Gaze redirection network training.} 
    Unlike previous approaches~\cite{Kononenko2017,Ganin2016}, our redirection network is pre-trained with synthetic
    eye images so that a large amount of well aligned image pairs (the same eye position, eye size, head pose and illumination)
    can be exploited.
    As a result, thanks to the large amount of data, the network does not require the eye landmarks as anchoring points.
  Besides, we also propose to exploit the segmentation map of synthetic samples for regularization during training.

  \item \textbf{Gaze redirection domain adaptation.} 
    Training with synthetic data results in the  domain shift problem.
    However, as we do not have aligned pairs of real images to do domain adaptation,
    we proposed instead a self-supervised method relying on a cycle consistency loss and a gaze redirection loss. 
  \item \textbf{Person-specific gaze adaptation using gaze-redirected samples.} 
    We hypothesize that these samples will provide more diverse visual content and  gaze groundtruth 
    compared to the reference samples they originated from, thus improving the person-specific gaze adaptation.
    To the best of our knowledge, we are the first to propose this idea and a series of experiments
    to validate its efficacy.
\end{compactitem}

\vspace*{1mm}

The rest of the paper is organized as follows. We first summarize the related works in Section 2 and then introduce our method in Section 3.
Experimental results are reported in Section 4 while a brief discussion is made in Section 5. The final conclusion is given in Section 6.

% - Gaze, human behaviour, how important
% - Problem of Gaze estimation, lack of data, personal bias
% - current solution, SimGAN, Wang's method, Gang's method 

% - Gaze redirection, usually applied in ....
% - advantage of Gaze redirection, domain difference, gaze bias, inspired by []...., suitable for personalized ...

% - Gaze redirection, learn from syntheis - why synthesis (aligned) - domain adaptation - how adapt

% - used for personalized gaze, .... user defined gaze difference ...

% - paper organization 

\section{Related Works}

\mypartitle{Gaze Estimation.}
As stated in introductionm, non-invasive vision based gaze estimation methods can be divided into geometric ones (GBM) and
appearance based ones (ABM)~\cite{hansen2010eye}.
GBMs build eye models based on some eye features, such as eye corners or iris localization and infer gaze direction using geometric relationship
between elements like the line joining the eyeball center to the iris center ~\cite{Wang2018,Wood2016,FunesMora2014,Wood2014,Wang2017,Wang2018a}.
Usually they do not require much training samples except for a few  calibration points,
but they suffer from low resolution imaging, noise and variable lighting conditions.

ABMs are more robust to those factors~\cite{Funes-Mora2016,Lu2014,Wood2015,Sugano2014,Lu2011a},
as they learn a regressor from annotated data samples and estimate gaze directly from the images.
%thus they are potentially more robust even dealing low resolution images
In particular, recently
deep learning approaches~\cite{Fischer2018,Zhang2018,Wang2018,Palmero2018,Zhang2017a,Zhang2015,Krafka2016}
have been shown to work well because
they train a regression network leveraging large amounts of data.
They can capture what are the image features essential  for gaze estimation
under various conditions, such as various eye shapes, illumination, glasses and head pose.

\mypartitle{Gaze Adaptation.}
However, when testing on unknown person, the different personal eye structures such as eye shapes and visual axis limit the performance of both GBMs and ABMs~\cite{Liu2018}. Some straightforward solutions to this problem have been proposed, such as to learn person-specific models~\cite{Linden2018,Zhang2015,Sugano2014}, fine-tune a pre-trained model~\cite{Masko2017}, learn a SVR using a few samples for calibration~\cite{Krafka2016} or learn a differential gaze model~\cite{Liu2018}.

Training a person-specific model or fine-tuning a pre-trained model can achieve very high accuracy for such person, but it usually requires relatively large amount of annotated data from this person, which is not wanted in practice. Calibrating person-specific model with an SVR or relying on differential gaze only require a few reference annotated samples, but those samples do not reflect the global gaze map, and the estimation error will increase when the gaze difference between the test sample and the reference sample becomes large.

Under this circumstance, we propose a gaze redirection method that can alleviate the drawbacks from the aforementioned methods. More precisely, our algorithm can generate more diverse and realistic images using a few annotated samples from this person. Then these data can be used to fine-tune a pre-trained gaze model.

\mypartitle{Gaze Redirection.} As far as we know, the computer vision and graphics based gaze redirection for video-conferencing was first studied in~\cite{Zitnick1999}, in which two components are included to solve this task. The first is tracking the user's head pose and eye ball motion, and the second consists of manipulating the head orientation and eye gaze.
Following this work, Weiner \textit{et.al.}~\cite{Weiner2002} evaluated and proved the overall feasibility of gaze redirection in face images via eye synthesis and replacement by integrating the vision and graphical algorithm within a demonstration program.
But changes in the eyelid configuration were not considered.
Then a simple solution that detects eyes and replaces them with eye images in a front gaze direction was proposed in~\cite{Wolf2010,Qin2015}.
Kononenko \textit{et.al.} proposed a pixel-wise replacement method using an \textit{eye flow tree} and could synthesize realistic views with a gaze systematically redirected upwards by 10 to 15 degrees~\cite{Kononenko2015}.
Then they updated the eye flow tree by a deep warping network trained on pairs of eye images corresponding to eye appearance before and after the redirection~\cite{Ganin2016,Kononenko2017}. However, these methods require large amount of annotated data for training. 

To circumvent this issue, Wood~\cite{Wood2018} proposed a model based method that does not need any training samples.
%More precisely,
It first builds and fits a multi-part eye region model using an analysis-by-synthesis method to simultaneously recover the eye region shape, texture, pose, and gaze
for a given image. Then, it manipulates the eyes by warping the eyelids and rendering eyeballs in the output image.
It achieves better results especially for large redirection angles. 

% Different from the existing methods, our algorithm ...

\vspace*{-1mm}

\section{Gaze Adaptation approach}

\begin{figure*}[tb]
  \centering
  \hspace{1em}\includegraphics[height=75mm]{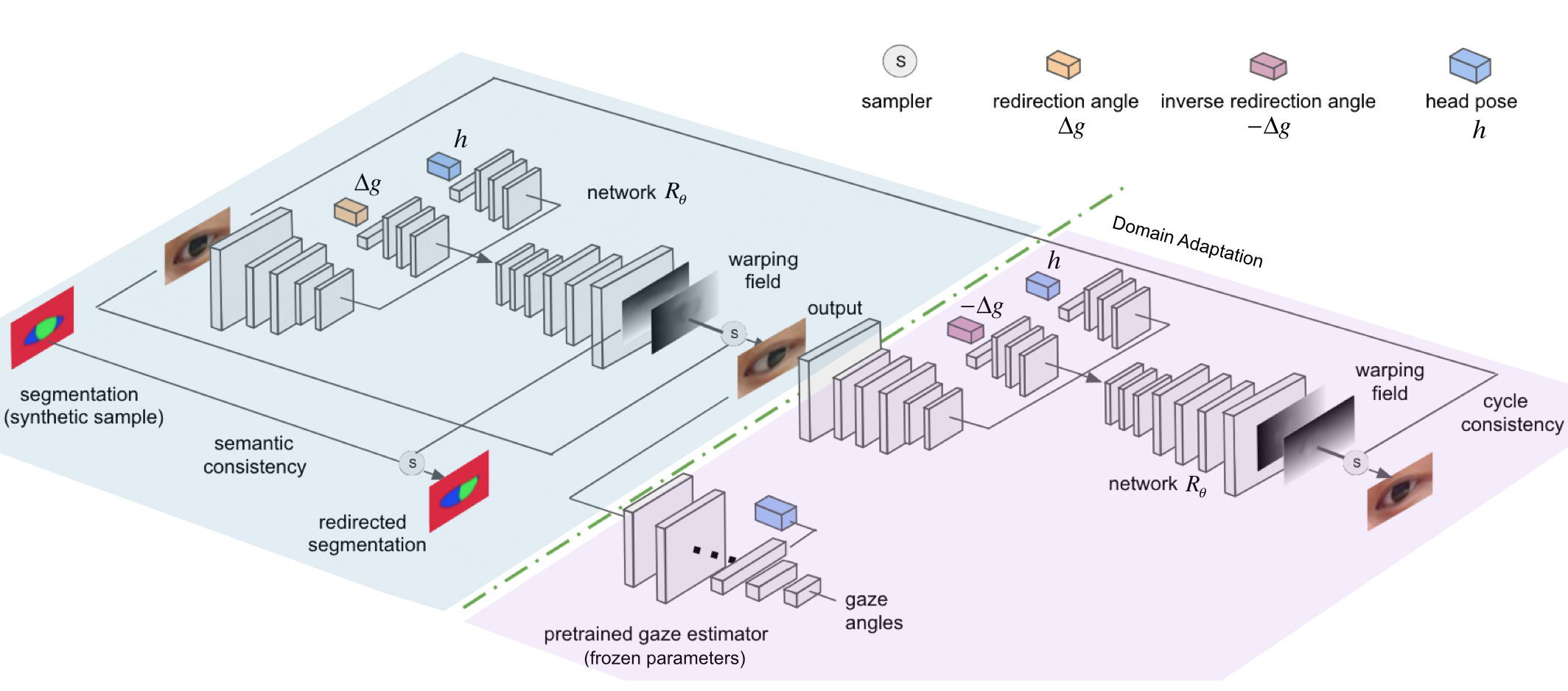}
  \vspace*{-1mm}
  \caption{Gaze redirection network (top left), along with learning components (eye segmentation for semantic consistency,
  cycle consistency, gaze prediction consistency).
  }
  \label{fig:framework}
\vspace*{-4mm}
\end{figure*}

\vspace*{-1mm}

Our overall approach for user-specific gaze adaptation is illustrated in Fig.~\ref{fig:main_idea}.
It consists in fine-tuning a generic neural network using labeled training samples.
However, rather than only using the very few (less than 10) reference samples,
we propose to generate additional samples using a gaze redirection model.
As this redirection model is the  main component of our approach, we describe it with more details
in the sections~\ref{sec:gareredirectionoverview} to~\ref{sec:adaptation}.
The gaze adaptation part is then described in section \ref{sec:finetuning}.

\subsection{Gaze Redirection Overview}
\label{sec:gareredirectionoverview}

Our framework for gaze redirection is shown in Fig.~\ref{fig:framework}. It is composed of the redirection network itself and a domain adaptation module. The left part of Fig.~\ref{fig:framework} illustrates the redirection network which takes the eye image, the user defined redirection angle and the head pose as input. It is designed as an encoder-decoder manner where the output of the decoder is an inverse warping field. The gaze-redirected sample is then generated by warping the input eye image with the predicted inverse warping field (via a differentiable sampler). The right part of Fig.~\ref{fig:framework} is the domain adaptation module which is conducted in a self-supervised way through a cycle consistency loss and a gaze redirection loss.

\subsection{Synthetic Data for Gaze Redirection Learning}

\begin{figure}[tb]
  \centering
  \includegraphics[height=22mm]{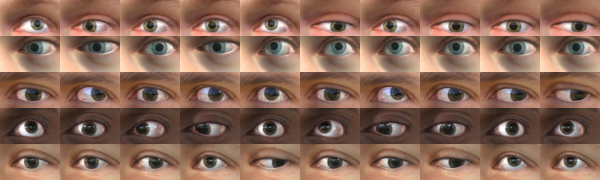}
  \vspace*{-1mm}
  \caption{Aligned UnityEyes samples (placed in rows)}
  \label{fig:unityEyes}
\vspace*{-2mm}
\end{figure}

In principle, the training of a gaze redirection network needs well aligned image pairs where the two images (the input one and the redirection groundtruth for supervision) share the same overall illumination condition, the same person-specific properties (skin color, eye shape, iris color, pupil color) and the same head pose. 
The only difference should be gaze-related features such as eye ball orientation and eyelid status. 
This strict requirement make it hard fto collect real data.
In this paper, we propose to use synthetic samples instead. 
Concretely, we use the UnityEyes Engine~\cite{wood2016learning} to produce 3K eye image groups, each containing 10 images generated with the same illumination, the same person-specific parameter, the same head pose, but different gaze parameters, as shown in Fig.~\ref{fig:unityEyes}.
A totol of 10*9 image pairs can thus be drawn from each group.
In our work, we used 10K image pairs for training.

% Concretely, we randomly sample 1K person-specific parameters (controlling properties like color of skin, iris and pupil) in UnityEyes engine. For each person-specific parameter we generate 30 frames where the geometry parameter (mainly the head pose) changes every 10 frames and the gaze changes every single frame. 
%
%
% By this way, we generated 30K samples which can be categorized into 3K groups. The samples in a group share the same properties except gaze. Therefore, we can draw 10*9 image pairs from each group and collect 3000*10*9 redirection pairs. In our work, we only use 10K pairs for training. 
% %
% We show some groups by rows in Fig.~\ref{fig:unityEyes}. 

\subsection{Gaze Redirection Network}

\mypartitle{Architecture.}
%<
It is illustrated in Fig.~\ref{fig:framework}. 
The network takes three variables as input, the eye image \img, the head pose \headPose and the user defined redirection angle \deltaGaze. Among them, \img is processed by an image branch and encoded as a semantic feature, while \headPose and \deltaGaze are processed with another two branches and encoded as features which will guide the gaze related visual changes.
Note that the head pose input is a must since it is one of the elements which determine the appearance of eye images. 
The three output features are then stacked in a bottleneck layer and further decoded into two inverse warping maps \warpX and \warpY:
\begin{equation}
\warpXY = \RNet_{\RNetParameter}(\img, \deltaGaze, \headPose)
\label{eq:mapGenerate}
\end{equation}
where \RNet is the redirection network and \RNetParameter is the network parameter.
Similarly to~\cite{Kononenko2017}, we then use a differentiable grid sampler \gridSampler~\cite{NIPS2015_5854} to warp the input image and generate the gaze-redirected image $\img_{\deltaGaze}$ whose gaze groundtruth is $\gazeLabel + \deltaGaze$ (\gazeLabel is the gaze of the original image \img) according to:
\begin{equation}
\setlength{\jot}{-10pt}%
\resizebox{0.9\hsize}{!}{$
\begin{aligned}
\img_{\deltaGaze}(x,y) = \sum_{i}\sum_{j}\img(i,j) &\cdot max(0,1-|i-\warpX(x,y)|)\\ 
&\vspace{-2em}\cdot max(0,1-|j-\warpY(x,y)|).
\end{aligned}$}
\label{eq:imgRedirection}
\end{equation}
%\vspace{0.3em}
%
For simplicity, we rewrite the above formulas as:
\begin{equation}
\img_{\deltaGaze}=\img \circ \RNet_{\RNetParameter}(\img, \deltaGaze, \headPose)
\label{eq:gridSample}
\end{equation}
where $\circ$ represents the warping operation.
Compared with direct synthesis, this strategy projects the pixels of the input to the output,
which guarantees that the input and the output will share similar color and illumination distributions.
%
% The advantage of this strategy over direct synthesis is that the warping field rearranges the pixel in the input to the output, which to some extent guarantees that the input and output share a similar distribution. This property is quite important when the testing domain of the model is different to the training domain.

For training, we use an L1 loss to measure the difference between the redirection output $\img_{\deltaGaze}$ and the groundtruth $\gtI$.
Therefore, generating the required inverse warping field for redirection is learned in an indirect supervised way.

\begin{figure}[tb]
  \centering
  \includegraphics[height=30mm]{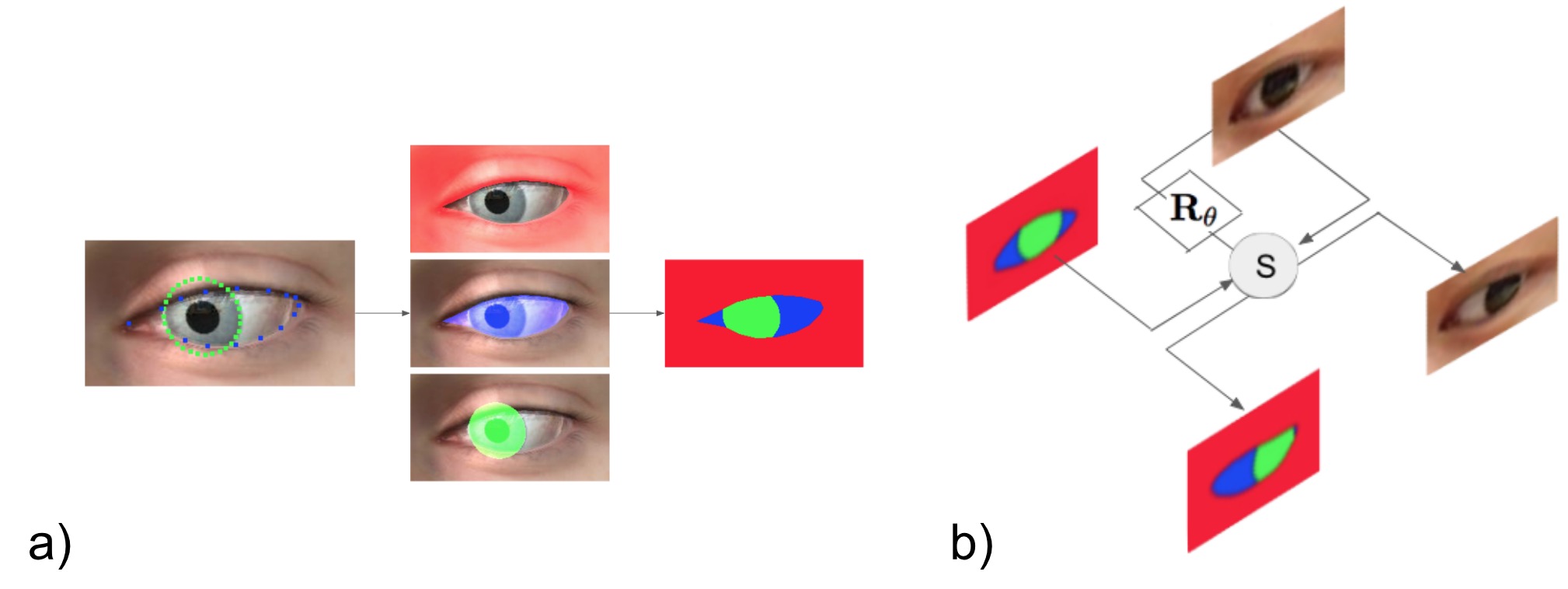}
  % \vspace*{-5mm}
  \caption{Semantic consistency. (a) Deterministic segmentation of a synthetic sample, red: background, blue: sclera, green: iris $+$ pupil.
 (b) the gaze redirection of a segmentation map.}
  \label{fig:seg}
\vspace*{-2mm}
\end{figure}

\mypartitle{Semantic consistency.} 
So far, the network can be evaluated by measuring the reconstruction loss between the predicted gaze-redirected eye image $\img$
and the corresponding groundtruth \gtI.
If the predicted inverse warping field is accurate, then the different semantic parts of the eye (pupil, sclera and background) should also be well redirected.
We thus propose to enforce the warping consistency at the semantic level.
To do so, for each synthetic image \img, we extract the semantic map as follows: we first fit convex shapes to the eyelid landmarks and the iris landmarks (provided by UnityEyes) to get the maps of the iris $+$ pupil region, the sclera region and the background region. We then merge these  three maps into a segmentation map \segI, as shown in Fig.~\ref{fig:seg}a.
It is important to note that this step is deterministic and is not a part of the network. 
%
% An eye image can be segmented into three semantic parts, the eyeball region, the sclera region and the background (including the skin and anything else), shown in the rightmost image in Fig.~\ref{fig:seg}. 
% %
% For a well predicted inverse warping field, the warping should not only generate accurate gaze-redirected images, but also precise gaze-redirected segmentation maps, shown in Fig.~\ref{fig:seg}b. It inspires us to propose a segmentation regularization which enforces consistency at semantic level. 
%
Then, any segmentation map \segI can then be redirected with the inverse warping field $\RNet_{\RNetParameter}(\img, \deltaGaze, \headPose)$
(which is predicted from the original image \img) and compared with the segmentation map \gtSeg of the target redirected eye \gtI.

\mypartitle{Overall loss.}
According to previous paragraphs, our overall redirection loss $L_R$ (for synthetic data) can be defined as the sum of a reconstruction loss
and of the semantic loss, using in each case L1 norms. It is thus defined as:
%
%We take a L1 loss on the redirected segmentation $\segI \circ \RNet_{\RNetParameter}(\img, \deltaGaze, \headPose)$ and the groundtruth segmentation \gtSeg. The total loss of the network is the combination of the image loss and the semantic regularization, shown in Eq.~\ref{eq:redirectionLoss}.
%
\begin{equation}
\resizebox{0.9\hsize}{!}{$
L_R = ||\img \circ \RNet_{\RNetParameter}(\img, \deltaGaze, \headPose)-\gtI||_{1} + ||\segI \circ \RNet_{\RNetParameter}(\img, \deltaGaze, \headPose)-\gtSeg||_{1}$}
\label{eq:redirectionLoss}
\end{equation}
Please note that the segmentation map is not processed by the network (looking at Fig.~\ref{fig:seg}b)
and will not be required at user gaze adaptation time for generating redirected samples.

% \begin{figure}[tb]
%   \centering
%   \includegraphics[height=25mm]{img/seg.png}
%   \vspace*{-1mm}
%   \caption{Segmentation of UnityEyes samples}
%   \label{fig:seg}
% \vspace*{-2mm}
% \end{figure}

\subsection{Gaze Redirection Domain Adaptation}
\label{sec:adaptation}

Because of the domain difference between synthetic and real data, the performance of the network $\RNet_{\RNetParameter}$ learned only from synthetic data degrades when it is applied to real data.
A straightforward solution to solve this issue would be to fine tune $\RNet_{\RNetParameter}$ with real image pairs.
However, as mentioned above, collecting real image pairs for gaze redirection is difficult.
In this section, we introduce a self-supervised domain adaptation method relying on two principles.
The first one is gaze redirection cycle consistency, and the second one is based on the consistency of the estimated gaze from the
gaze redirected image.

\mypartitle{Cycle consistency loss.}
It has been used for applications like domain adaptation~\cite{DBLPZhuPIE17} and identity preserving~\cite{pumarola2018ganimation}.
The main idea is that when a sample is transferred to a new domain and then converted back to the original domain, the cycle output should be the same as the input.
Similarly, in our case, if a gaze redirected sample $\img_{\deltaGaze}$ is further redirected with the inverse redirection angle -\deltaGaze, the cycle output should be close to the original image \img. 

In this paper, we apply this cycle consistency scheme to the set of real images, and define the cycle loss as:
\begin{equation}
\begin{aligned}
L_{cycle} = ||\img_{\deltaGaze} \circ \RNet_{\theta}(\img_{\deltaGaze}, -\deltaGaze, \headPose)-\img||_{1} 
\end{aligned}
\label{eq:cycleLabelAdpt}
\end{equation}
where $\img_{\deltaGaze}=\img \circ \RNet_{\theta}(\img, \deltaGaze, \headPose)$.

\mypartitle{Gaze redirection loss.}  
As a weakness, the cycle loss alone could push the redirection network to collapse to an identity mapping (the output of the redirection network is always equal to the input).
To prevent this collapse, we propose to exploit a gaze redirection loss. 
More concretely, given a set of real data, we first train a generic gaze estimator \gazeNet using them.
We then freeze the parameters of \gazeNet and use it to define a loss on the gaze-redirected image,
enforcing that the gaze predicted from this image should be close to its target groundtruth (see bottom of Fig.~\ref{fig:framework}).
More formally:
\begin{equation}
L_{gaze} = ||\gazeNet(\img \circ \RNet_{\theta}(\img, \deltaGaze, \headPose))-(\gazeLabel + \deltaGaze)||_{2}
\label{eq:labelAdpt}
\end{equation}
Besides preventing the collapse, the real data trained gaze estimator \gazeNet can help reducing the systematic bias in the gaze redirection network
(arising from intially training the network with only synthetic data) and therefore help the domain adaptation of $\RNet_{\theta}$.

\mypartitle{Network adaptation optimization.} 
To conduct network adaptation, we do not consider the two losses in the same minibatches, as they are of different nature.
In addition, to balance the domain adaptation and the gaze redirection, not all parts of the network need to be adapted simultaneously.
In practice, we thus optimize the two losses alternatively according to the following scheme.
For the cycle loss $L_{cycle}$, we only optimize the image encoding branch since
i) domain shift usually occurs when encoding an input image into semantic features;
ii) the fixed decoder part can further prevent the redirection network from collapsing.
For the gaze redirection loss $L_{gaze}$, only the head pose and gaze branches are updated.
The image encoder and decoder remain frozen in this case to prevent an overfitting to $L_{gaze}$.
We use Stochastic Gradient Descent (SGD) to optimize the network.

\subsection{Gaze Estimation Adaptation}
\label{sec:finetuning}

As stated earlier, the aim of the gaze redirection is to generate more person-specific samples for gaze adaptation.
In our work, we first train a generic gaze estimator using the real data from several identities.
We then adapt the estimator with the samples of a new person and their gaze-redirected outputs.
This adaptation is conducted in a few-shot setting, meaning the number of original samples of this new person is few (less than 10). 
%
%In the case of a network based gaze estimator, we use network fine tuning for gaze adaptation.
%
More concretely, the generic estimation network is fine tuned with the person-specific samples during 10 epochs.
In the first 5 ones,  we use both the original  and the gaze-redirected samples,
while in the last 5 ones we only use the original samples to minimize the effects
of potentially wrong  redirected samples.
Since the number of samples is small, we use Batch Gradient Descent instead of Stochastic Gradient Descent.
Further details about the generic gaze estimator and its adaptation can be found in the Experiment Section.

\section{Experiment}

In experiments, our main aim is to evaluate the performance of the person-specific gaze estimators adapted from a generic estimator using few reference samples 
%
%, we mainly measure the performance of the person-specific gaze estimator which is adapted from a generic gaze estimator using the few-shot person-specific samples
%
and their gaze-redirected samples.
Nevertheless, we also conduct a subjective test to evaluate to which extent
the redirected samples are realistic enough for humans.
Note that in this paper, we only target single eye image gaze estmtion (and redirection and adaptation),
leaving the full-face case as future work.

\subsection{Experimental Setting}

\begin{figure*}[tb]
 \centering
%\subfigure[ColumbiaGaze samples]{
a) \includegraphics[height=70mm]{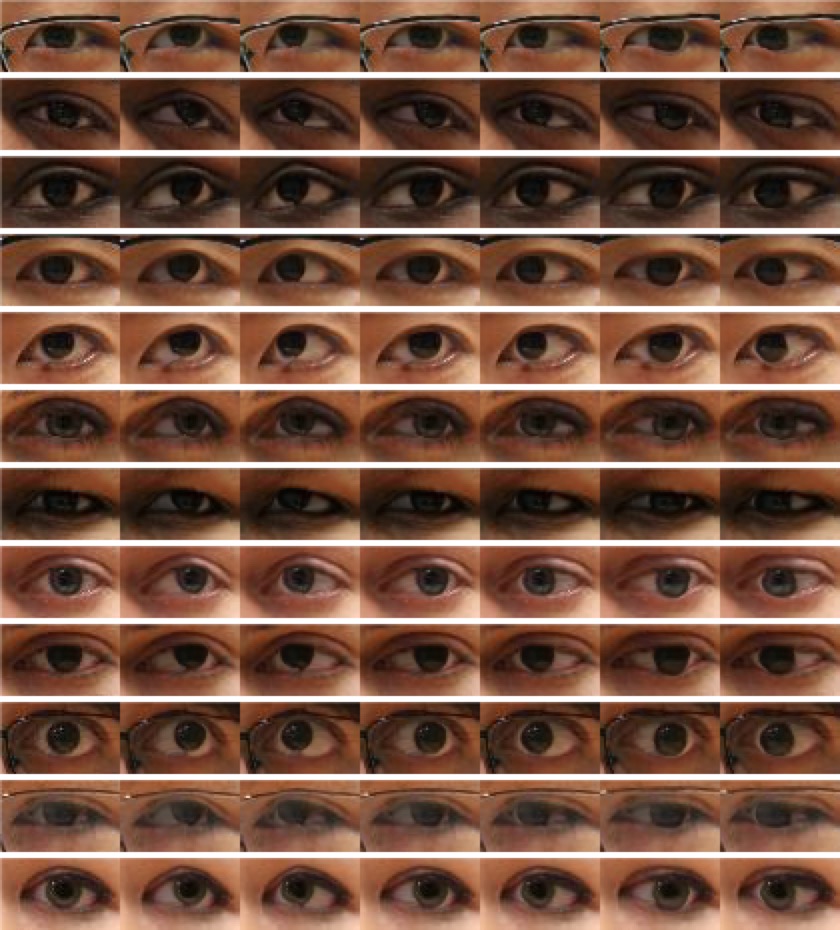} 
% }
%\subfigure[MPIIGaze samples]{
\hspace{1.5em}
b) \includegraphics[height=70mm]{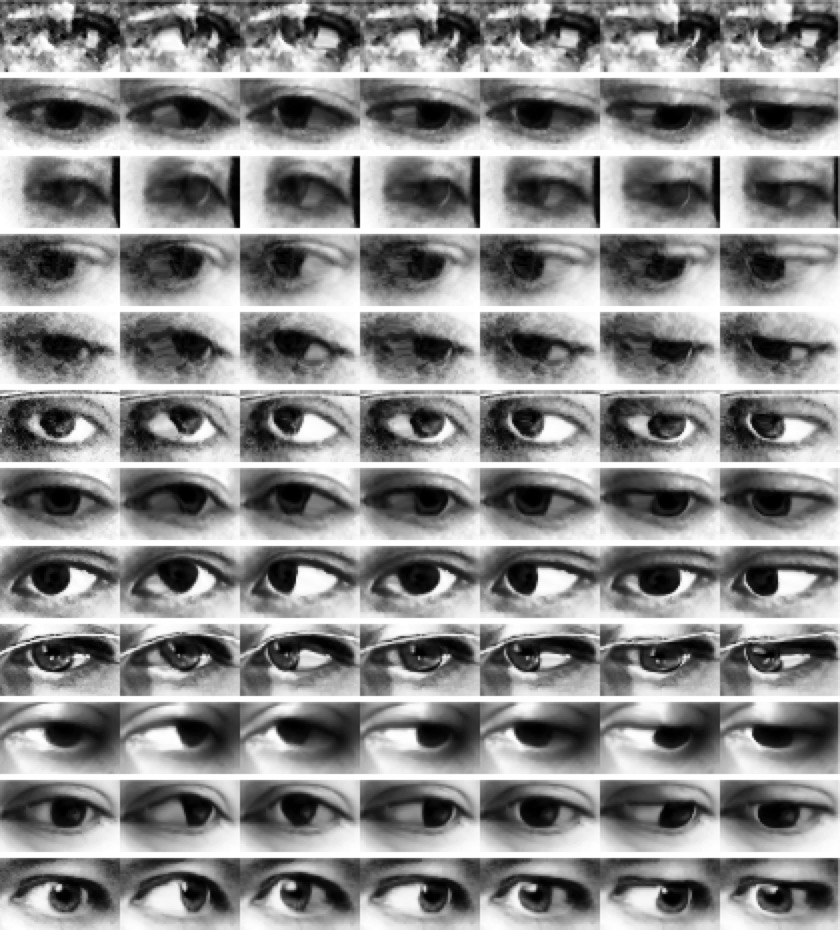}
%}
%\subfigure[Pairs of subjective test]{
  %
\hspace{1.5em}
c) \includegraphics[height=70mm]{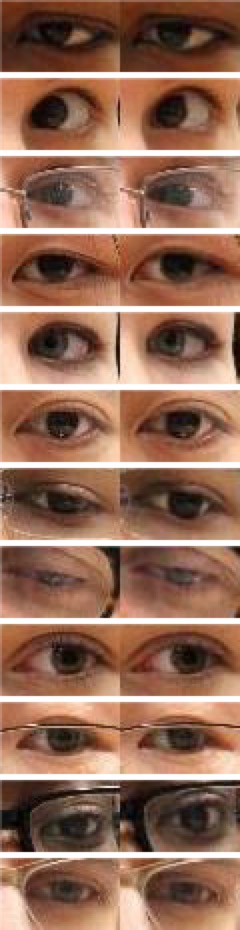}
\hspace{1.5em}
% }
\vspace*{-1mm}
\caption{Redirection qualitative results from the ColumbiaGaze (a) and MPIIGaze (b) datasets.
  In (a) and (b), the first image of each row is an original sample, whereas the remaining images in the row are redirected samples from this original sample.
  Subfigure (c) displays pairs of images used in the subjective test: in each pair, the left image is an original image from the dataset,
  while the right one is a redirected sample (obtained from another original sample) which has the same gaze label (i.e. direction) as the left one.
 }
\label{fig:subjective_test}
\vspace*{-2mm}
\end{figure*}

\mypartitle{Datasets.} We use the ColumbiaGaze Dataset~\cite{smith2013gaze} and the MPIIGaze Dataset~\cite{Zhang2015} for experiment.
The former one contains the gaze samples of 56 persons while the latter contains eye images of 15 persons.

\mypartitle{Generic gaze estimator.}
As our gaze estimator, we use GazeNet~\cite{Zhang2017a}.
It is is based on a \vgg architecture.
To train it, we follow the protocols of the ColumbiaGaze and MPIIGaze datasets (i.e. as for cross-subject experiments),
using respectively a  5-fold and 15-fold training scheme.
The error of our generic gaze estimator on ColumbiaGaze is $3.54^{\circ}$ ($3.9^{\circ}$ in~\cite{Park2018a}) while the error on MPIIGaze is $5.35^{\circ}$ ($5.5^{\circ}$ in~\cite{Zhang2017a}), showing better performance than the state-of-the-art results. 
Please note that the generic gaze estimator\footnote{A generic estimator is trained for each fold.
  In none of the experiments, data from the test subject is used in either part of the training phase.}
is also exploited as \gazeNet to define the gaze redirection loss, as defined in section~\ref{sec:adaptation}.

\mypartitle{Evaluated models.}
Starting from the generic gaze estimator, we develop a series of adaptation methods to  contrast with our approach.
The first two methods are the linear  ($\LinAdpt$, \cite{Liu2018}) and
the SVR  ($\SVRAdpt$, using the features of the second last layer~\cite{Liu2018, Krafka2016}) gaze adaptation methods
which learn  additional regressors from the gaze estimator output ($\LinAdpt$) or features ($\SVRAdpt$),
and thus do not change (or adapt) the generic gaze estimator.
%
% $\LinAdpt$ (linear gaze adaptation~\cite{Liu2018}) and 
%$\SVRAdpt$ (SVR gaze adaptation using the features of the second last layer~\cite{Liu2018, Krafka2016})
% which learn additional regressors and do not change the generic gaze estimator. 
%
In contrast, the third and fourth approaches directly fine tune the generic estimator using either only the reference samples ($\Finetune$, $\FT$ for fine tuning) or
as well the gaze redirected samples ($\AugFinetune$, $\Red$ for redirection).
%
%We name the \textbf{F}ine \textbf{T}uning with original samples as $\Finetune$ and the \textbf{F}ine \textbf{T}uning with both original samples and \textbf{Red}irected samples as $\AugFinetune$.
%

In addition, we also implement a differential gaze estimator \difNet~\cite{Liu2018} for comparison.
The \difNet is trained to predict gaze differences, and it exploits the reference samples
to predict the gaze of a new eye image.
For a fair comparison, we replace the three convolution layers used as feature extractor in~\cite{Liu2018} with the  \vgg feature extractor. 
%
%is inspired by Siamese network and is first proposed in~\cite{Liu2018} where the network has only three convolution layers. For a fair comparison, we replace the feature extraction layers in~\cite{Liu2018} with the \vgg feature extractor.
Please note that the \difNet approach can be regarded as a person-specifc network since person-specific samples (at least one)
are required to  estimating the gaze of  new eye image.

\mypartitle{Gaze redirection parameters.} 
For each person, we randomly draw \numSample (\numSample $=$1, 5 or 9) person-specific samples
and generate $\redTime \cdot \numSample$ gaze-redirected samples where the default value of  \redTime is 10.
For the MPIIGaze dataset in which the gaze groundtruth is continuous, the yaw and pitch components $(\deltaGaze_{p},\deltaGaze_{y})$
of the redirection angle $\deltaGaze$ are randomly chosen with the range $[-10,10] \times [-15,15]$ ($[-10,10]$ for pitch,
and $[-15,15]$ for yaw).
For the ColumbiaGaze dataset, where the annotated gaze is discrete, \deltaGaze is chosen from the same range
but with discrete values ($\pm5^{\circ}, \pm10^{\circ}, \pm15^{\circ}$).
The impact of $\redTime$ and of the redirection ranges are  further studied in the result section. 

\mypartitle{Performance measurement.} 
We use the angle (in degree) between the predicted gaze vector and the groundtruth gaze vector as the error measurement.
Note that gaze vectors are  3D unit  vectors constructed from  the pitch  and  yaw angles.
To eliminate  random factors, we performed 10 rounds of person-specific sample selection,
gaze redirection and gaze adaptation, and reported the average estimation error.

% We use the redirection network described above to generate the gaze-redirected samples. To eliminate the random factors, 

% we did 10 rounds of evaluation.
% %
% In each round, we randomly draw \numSample user-specific reference samples and generate $\redTime \cdot \numSample$ redirected images for network adaptation. The reported results are the average error of the 10 rounds.
% %
% In our experiment, we choose \numSample as 1, 5 and 9 respectively for evaluation. The default value of redirection times $\redTime$ is 10. But the impact of $\redTime$ would be further analyzed in the experiment.

% For MPIIGaze where the gaze labels are continuous, the redirection angle \deltaGaze is randomly chosen from the range $[-10,10] \times [-15,15]$ ($[-10,10]$ for pitch angle $\deltaGaze_{p}$ while $[-15,15]$ for yaw angle $\deltaGaze_{y}$). For ColumbiaGaze which annotated discrete gaze labels, \deltaGaze is chosen from a discrete sets with the same range. The impact of different redirection ranges would also be studied in the result section.

\subsection{Results}

\mypartitle{Gaze redirection qualitative results.}
We show some qualitative results of the redirection network in Fig.~\ref{fig:subjective_test}(a) and (b).
As can be seen, our redirection network does a realistic synthesis for samples with different skin or iris color.
Furthermore, we also found that the redirection model is  robust when working with  noisy eye images, as illustrated in several rows
of Fig.~\ref{fig:subjective_test}(b).

\begin{table}[tb]\scriptsize
 \caption{ColumbiaGaze dataset: gaze adaptation performance}
\centering
\begin{tabular}[tb]
{@{}p{2.0cm}|p{0.7cm}p{0.6cm}p{0.7cm}p{0.7cm}p{0.45cm}p{1.3cm}@{}}
\hline
\diagbox{\#sample}{error}{approach} & \textit{Cross } 
\newline \textit{Subject} & \multirow{2}{*}{$\LinAdpt$} & \multirow{2}{*}{$\SVRAdpt$} & \multirow{2}{*}{$\Finetune$} & \multirow{2}{*}{$\difNet$} & \multirow{2}{*}{$\AugFinetune$}\\
\hline
\hspace{4em}1 & \multirow{3}{*}{3.54} & \hspace{0.4em}- & \hspace{0.4em}- & 5.53 & 4.64 & \textbf{3.92}\\
\hspace{4em}5 & & 4.65 & 7.67 & 3.11 & 3.63 & \textbf{2.88} \\
\hspace{4em}9 & & 3.78 & 5.39 & 2.79 & 3.50 & \textbf{2.60} \\
\hline
\end{tabular}
\label{tab:model_comp_columbia}
\end{table}

\begin{table}[tb]\scriptsize
 \caption{MPIIGaze dataset: gaze adaptation performance}
\centering
\begin{tabular}[tb]
{@{}p{2.0cm}|p{0.7cm}p{0.6cm}p{0.7cm}p{0.7cm}p{0.45cm}p{1.3cm}@{}}
\hline
\diagbox{\#sample}{error}{approach} & \textit{Cross } 
\newline \textit{Subject} & \multirow{2}{*}{$\LinAdpt$} & \multirow{2}{*}{$\SVRAdpt$} & \multirow{2}{*}{$\Finetune$} & \multirow{2}{*}{$\difNet$} & \multirow{2}{*}{$\AugFinetune$}\\
\hline
\hspace{4em}1 & \multirow{3}{*}{5.35} & \hspace{0.4em}- & \hspace{0.4em}- & 5.28 & 5.93 & \textbf{4.97}\\
\hspace{4em}5 & & 5.43 & 7.68 & 4.64 & 4.42 & \textbf{4.20} \\
\hspace{4em}9 & & 4.61 & 5.79 & 4.31 & 4.20 & \textbf{4.01} \\
\hline
\end{tabular}
\label{tab:model_comp_mpii}
\end{table}

\mypartitle{Gaze adaptation performance.}
They are reported in  Table.~\ref{tab:model_comp_columbia} (ColumbiaGaze dataset) and Table.~\ref{tab:model_comp_mpii} (MPIIGaze dataset).
From the tables, we observe that the proposed approach $\AugFinetune$ achieves the best results while the $\LinAdpt$ and  $\SVRAdpt$ methods
obtain the worst results,  sometimes even degrading the generic gaze estimator. 
%¨%
The unsatisfactory performance of the latter models ($\LinAdpt$ and $\SVRAdpt$)
is probably due to the fact that the linear and SVR regressor do not make changes to the generic gaze estimator and
thus the capacity of gaze adaptation is limited.
%
%on adapt on predicted labels or feature while remain the gaze estimator freezed. 
%
We also find that the $\difNet$ is not always superior to the simpler $\Finetune$ approach.
This is surprising and shows that the ability of direct network fine tuning with small amount of data (less than 10) is often  overlooked
in the literature and  not even unattempted.
To the  best of our knowledge, we are the first to report this result which can inspire new research on user-specific gaze estimation.

When comparing $\AugFinetune$ with the best results of $\difNet$ and $\Finetune$, we note that our approach leads the performance by around $0.2^{\circ}$.
While this may seem a marginal improvement, a more detailed analysis of the results shows that our approach improves
the results of \textbf{84.2\%} of the subjects from the ColumbiaGaze dataset and of \textbf{80\%} of the subjects from the MPIIGaze dataset
(compared with the best results of {\it both} $\difNet$ and $\Finetune$),
which means that the improvements brought by $\AugFinetune$ are  stable and rather systematic. 

From the two tables, we  note that the performances of all the methods improve as the number of reference samples increases.
We can  also notice that our approach seems to have a larger advantage when the number of reference samples is small, 
demonstrating that the diversity introduced by our redirected samples is more important
when fewer person-specific gaze information is provided.

Finally, while in general adaptation methods improve results, we  observe on the ColumbiaGaze dataset that they all perform worse than the generic estimator
(cross-subject result)  when using only one reference sample.
This  is most probably due to the large variance of the head pose in this dataset,
which makes it difficult to learn (through adaptation) person-specific characteristics from only one sample.

\begin{table}[tb]\footnotesize
 \caption{ColumbiaGaze: Results with different redirection range}
\centering
\begin{tabular}[tb]
{@{}p{1.6cm}|p{0.9cm}p{1cm}p{1cm}@{}}
\hline
\diagbox{$\deltaGaze_{p}$}{error}{$\deltaGaze_{y}$} & $[-5,5]$ & $[-10,10]$ & $[-15,15]$ \\
\hline
$[-10,10]$ & 2.66 & 2.62 & \textbf{2.60}\\
\hline
\end{tabular}
\label{tab:redirect_range_columbia}
\end{table}

\begin{table}[tb]\footnotesize
 \caption{MPIIGaze: Results with different redirection range}
\centering
\begin{tabular}[tb]
{@{}p{1.6cm}|p{0.9cm}p{1cm}p{1cm}@{}}
\hline
\diagbox{$\deltaGaze_{p}$}{error}{$\deltaGaze_{y}$} & $[-5,5]$ & $[-10,10]$ & $[-15,15]$ \\
\hline
$[-5,5]$ & 4.15 & 4.06 & 4.02\\
$[-10,10]$ & 4.10 & 4.03 & \textbf{4.01}\\
\hline
\end{tabular}
\label{tab:redirect_range_mpii}
\vspace{-2mm}
\end{table}

\mypartitle{Redirection range.}
We use different gaze redirection ranges to generate samples for gaze adaptation. The selected redirection ranges are shown in Table.~\ref{tab:redirect_range_columbia} and Table.~\ref{tab:redirect_range_mpii}.
Note that we only use one redirection range of pitch for the ColumbiaGaze dataset since the gaze groundtruth in this dataset is discrete and there are only three values for the pitch angle, $-10^{\circ}, 0^{\circ}, 10^{\circ}$. It is thus not necessary to produce samples with new groundtruth.
From the results,  we find that larger redirection ranges do bring an improvement,
especially for the MPIIGaze dataset where the performance improves from $4.15^{\circ}$ to $4.01^{\circ}$.
This result is expected since a larger redirection range will usually bring more gaze diversity,
provided that the redirection module produces synthesized samples realistic enough for the given user.
Besides, we also find from Table.~\ref{tab:redirect_range_mpii} that a larger redirection range  for the  yaw angle
seems to be more effective than a larger redirection range for the pitch.

\begin{figure}[tb]
  \centering
  \hspace{1em}\includegraphics[height=45mm]{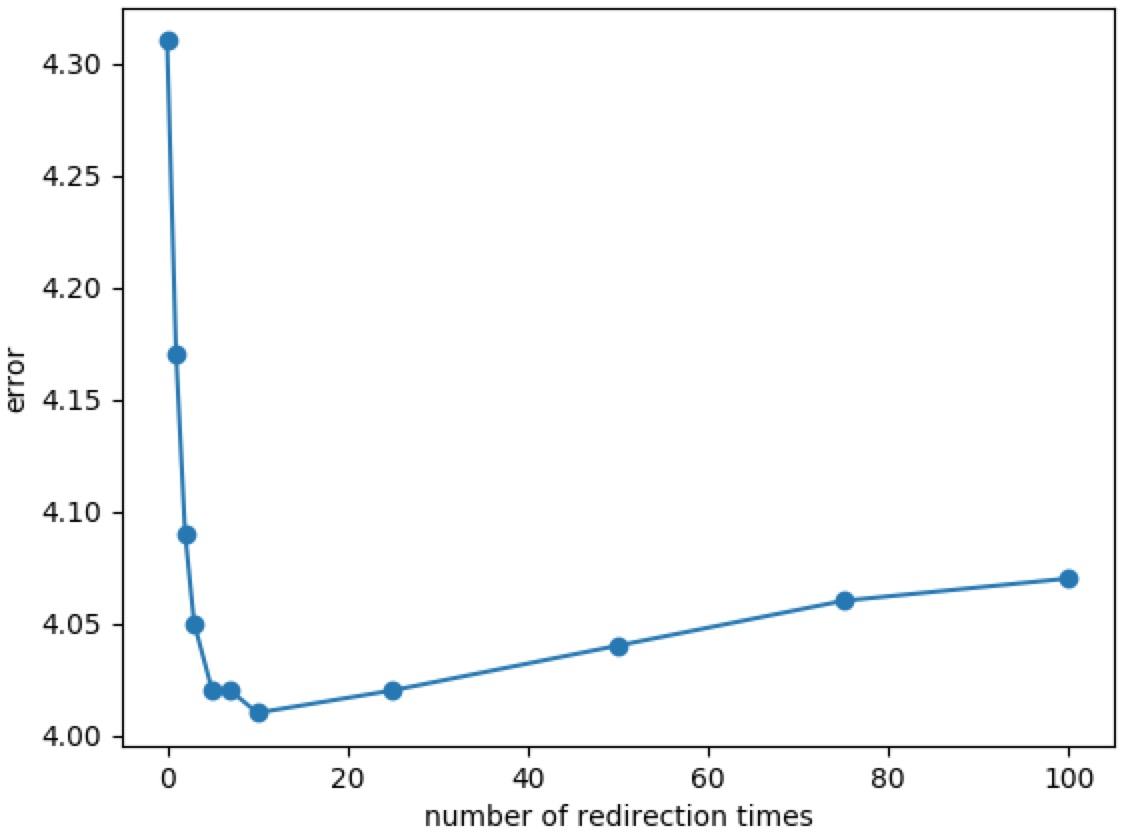}
  \vspace*{-1mm}
  \caption{Gaze adaptation performances w.r.t redirection times $\redTime$.}
  \label{fig:acc_red_num}
\vspace*{-2mm}
\end{figure}

\mypartitle{Number $\redTime$ of redirected gaze samples per reference sample.}
To study the impact of this parameter (the default value was  10 in all other experiments),
we randomly selected  9  reference samples for each person and generated $9\cdot\redTime$ gaze redirected samples,
varying $\redTime$ between $0$ and $100$.
We then adapted the generic gaze estimator with these samples as in all other experiments.
The corresponding performances are plotted in Fig.~\ref{fig:acc_red_num} for the MPIIGaze dataset (note that we do not use
the ColumbiaGaze dataset since its groundtruth and redirection angles are discrete, 
which limits the number of generated data).

The curve in Fig.~\ref{fig:acc_red_num} starts from $\redTime=0$ (which means only the initial reference   samples are used for adaptation).
As can be seen, the error decreases rapidly at first when $\redTime \in [0,5]$,
remains at a relatively  stable point within the range $\redTime \in [5,25]$, and then 
progressively degrades beyond that.
This curve shows that when $\redTime \simeq 10$, the generated samples provide enough diversity to adapt the network,
whereas beyond that, the use of too many samples results in an overfit of the network to the generated data which might not
reflect the actual distribution of eye gaze appearence of the user.

% \begin{table}[tb]\footnotesize
%  \caption{ColumbiaGaze: abilation study}
% \centering
% \begin{tabular}[tb]
% {@{}p{2.3cm}|p{1.6cm}p{1.6cm}p{1.4cm}@{}}
% \hline
% \diagbox{\#sample}{error}{approach} & $\AugFinetune$ \newline no adaptation & $\AugFinetune$ \newline no cycle & \multirow{2}{*}{$\AugFinetune$} \\ 
% \hline
% \hspace{4em}1 & 4.35 & 3.95 & \textbf{3.92} \\
% \hspace{4em}5 & 3.01 & 2.90 & \textbf{2.88} \\
% \hspace{4em}9 & 2.73 & 2.66 & \textbf{2.60} \\
% \hline
% \end{tabular}
% \label{fig:columbia_abilation}
% \end{table}

% \begin{table}[tb]\footnotesize
%  \caption{MPIIGaze: abilation study}
% \centering
% \begin{tabular}[tb]
% {@{}p{2.3cm}|p{1.6cm}p{1.6cm}p{1.4cm}@{}}
% \hline
% \diagbox{\#sample}{error}{approach} & $\AugFinetune$ \newline no adaptation & $\AugFinetune$ \newline no cycle & \multirow{2}{*}{$\AugFinetune$} \\ 
% \hline
% \hspace{4em}1 & 4.99 & 4.97 & \textbf{4.97} \\
% \hspace{4em}5 & 4.22 & 4.27 & \textbf{4.20} \\
% \hspace{4em}9 & 4.04 & 4.05 & \textbf{4.01} \\
% \hline
% \end{tabular}
% \label{fig:mpii_abilation}
% \end{table}

% \begin{figure}[tb]
%   \centering
%   \hspace{1em}\includegraphics[height=18mm]{img/cycle_contrast.png}
%   \vspace*{-1mm}
%   \caption{Contrast samples with and without cycle loss.}
%   \label{fig:cycle_contrast}
% \vspace*{-2mm}
% \end{figure}

\begin{table}[tb]\footnotesize
 \caption{Impact of the gaze redirection network domain adaptation (ColumbiaGaze dataset).}
\centering
\begin{tabular}[tb]
{@{}p{2.3cm}|p{1.1cm}p{2.1cm}p{1.4cm}@{}}
\hline
\diagbox{\#sample}{error}{approach} & $\Finetune$ & $\AugFinetuneNoDA$ & $\AugFinetune$ \\ 
\hline
\hspace{4em}1 & 5.53 & 4.35 & \textbf{3.92} \\
\hspace{4em}5 & 3.11 & 3.01 & \textbf{2.88} \\
\hspace{4em}9 & 2.79 & 2.73 & \textbf{2.60} \\
\hline
\end{tabular}
\label{fig:columbia_abilation}
\vspace*{-2mm}
\end{table}

\begin{table}[tb]\footnotesize
 \caption{Impact of the gaze redirection network domain adaptation (MPIIGaze dataset).}
\centering
\begin{tabular}[tb]
{@{}p{2.3cm}|p{1.1cm}p{2.1cm}p{1.4cm}@{}}
\hline
\diagbox{\#sample}{error}{approach} & $\Finetune$ & $\AugFinetuneNoDA$ & $\AugFinetune$ \\ 
\hline
\hspace{4em}1 & 5.28 & 4.99 & \textbf{4.97} \\
\hspace{4em}5 & 4.64 & 4.22 & \textbf{4.20} \\
\hspace{4em}9 & 4.31 & 4.04 & \textbf{4.01} \\
\hline
\end{tabular}
\label{fig:mpii_abilation}
\vspace*{-2mm}
\end{table}

\mypartitle{Domain adaptation.}
We remove the whole \textbf{D}omain \textbf{A}daptation step from the redirection network
%(i.e. it is only trained from synthetic samples)
and report the corresponding gaze adaptation results ($\AugFinetuneNoDA$) in Table.~\ref{fig:columbia_abilation} and Table.~\ref{fig:mpii_abilation}. 
On one hand, surprisingly, we note that exploiting the redirection network learned only from  synthetic
data still helps improving the gaze adaptation process ($\Finetune$ vs $\AugFinetuneNoDA$). 
On the other hand, when comparing $\AugFinetuneNoDA$ and $\AugFinetune$, we find that the domain adaptation further improves the gaze adaptation results.
This is particularly the case for the  ColumbiaGaze dataset.
A possible reason why the domain adaptation is less usefull  on the MPIIGaze dataset
is that the domain difference between MPIIGaze and the synthetic data
(all processed with histogram equalization to match MPIIGaze) is comparatively smaller.

\begin{figure}[tb]
 \centering
\hspace{-1.2em}
a){\includegraphics[width=40mm]{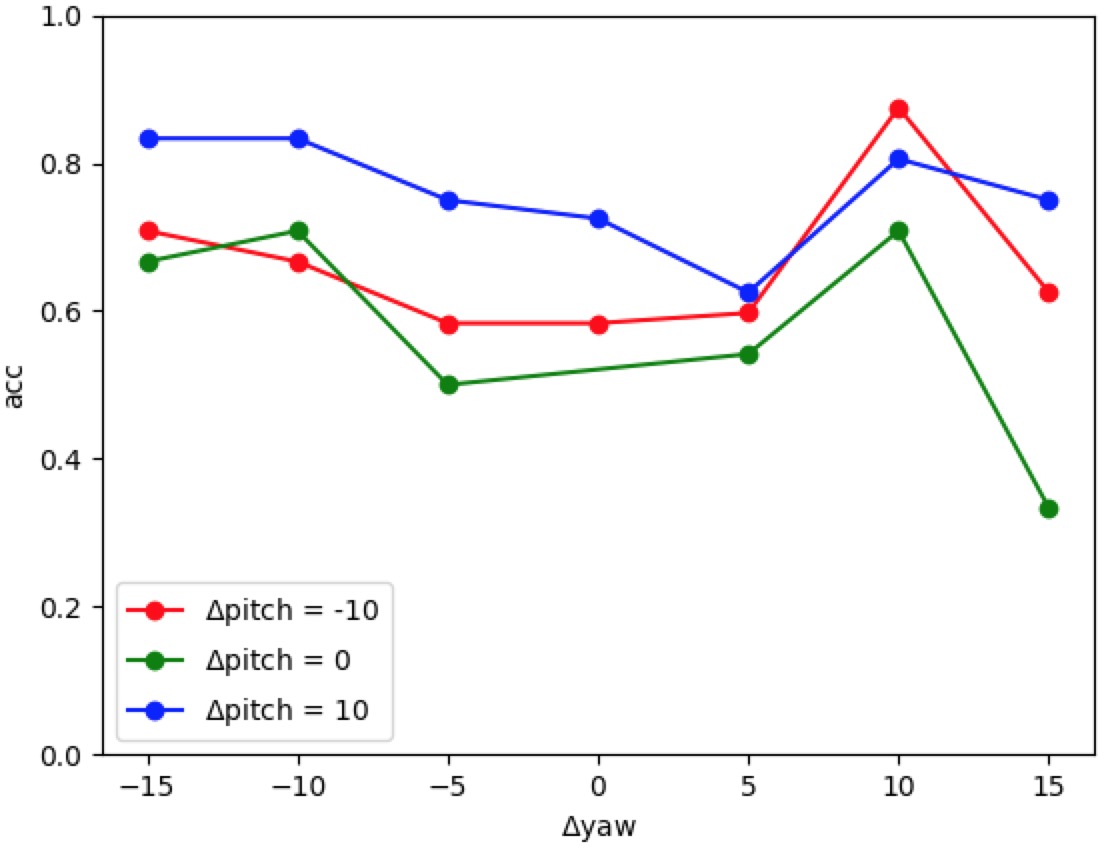} }
\hspace{-0.9em}
b){\includegraphics[width=40mm]{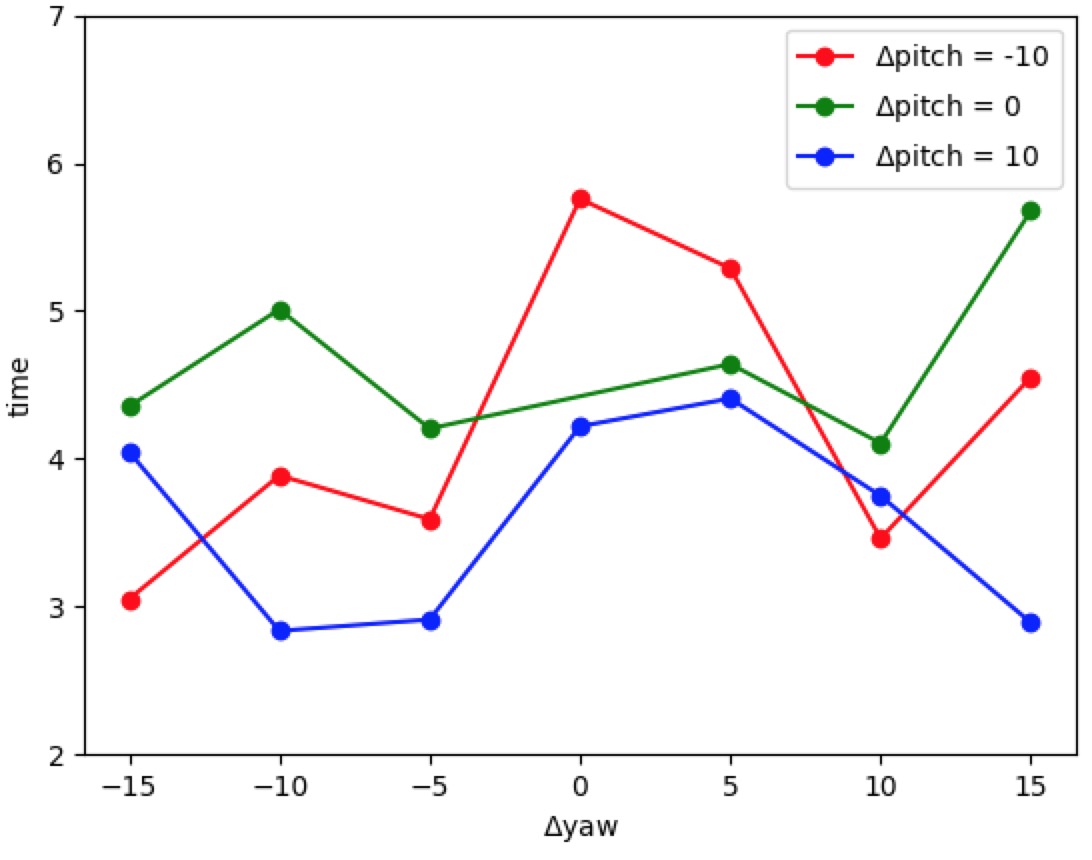}}
  \caption{Subjective test. (a) decision accuracy w.r.t redirection angles. (b) decision time w.r.t redirection angles.}
  \label{fig:subjective_test0}
\vspace*{-5mm}
\end{figure}

\mypartitle{Subjective test.}
To evaluate whether the gaze redirected samples are realistic, we invited 24 participants for a subjective test.
During the test, participants were shown 50 pairs of ColumbiaGaze samples,
where one image of the pair did correspond to an actual  real data sample,
and the second one was a gaze redirected sample.
Note that as a result, the eyes in each image pair share the same identity, the same gaze and the same head pose.
Some pairs are illustrated in Fig.~\ref{fig:subjective_test}c where the real images are all placed on the left for the purpose of demonstration.
In the test, the places of the real and redirected images were selected at random.
Participants were asked to choose the sample which they think was real.
A software was recording their choices as well as the time they took to make the decisions. 

Results are as follows.
The average accuracy of making a correct choice is 66\%, showing that distingusing genuine samples from redirected ones is difficult.
This is further confirmed by the average time to reach  a decision, which  is around  4 seconds and shows that 
people have to take some time to make a careful decision.

We also plot the decision precision and the decision time w.r.t redirection angles in Fig.~\ref{fig:subjective_test0}.
From Fig.~\ref{fig:subjective_test0}a, we find a general and expected trend that comparing samples
with  smaller redirection angles leads to more confusion, i.e. a low accuracy 
(and although as an artefact, the accuracy declines when $\Delta yaw = 15$).
The same  trend is observed in Fig.~\ref{fig:subjective_test0}b, where a smaller redirection angle corresponds to a longer decision time.
Nevertheless, in general, more participants and samples should be used to confirm these results, which we leave as a future work.

\section{Discussion}

In this section, we discuss techniques we attempted when developing the approach.

\mypartitle{More realistic redirected samples.}
Ganin et al.~\cite{Ganin2016} used a lighness correction refinement module 
on the gaze image redirected from the inverse warping field to produce a more realistic final redirected image.
It indeed removed a lot of artifacts in our case.
However, we found out that it was also degrading the performance of gaze adaptation,
because the refinement through a set of convolutional layers was altering too much
the distribution of color and illumination.

\mypartitle{GAN.}
We also attempted to use GAN (or CycleGAN when combined with the cycle loss) for domain adaptation.
However, as our redirected images are already of high quality,
the GAN did not further improve the gaze adaptation step.

\section{Conclusion}

We proposed to improve the adaptation of a generic
gaze estimator to a specific person from few shot samples via gaze redirection synthesis.
To do so, we first designed a redirection network that was pretrained from large amounts of well aligned synthetic data,
making it possible to predict accurate inverse warping fields.
We then  proposed a self-supervised method to adapt this model to real data.
Finally, for the first time to the best of our  knowledge, we exploited the gaze redirected samples to improve the performance of
a person-specific  gaze estimator.
Along this way, as a minor contribution, we also showed that the  simple fine tuning of a generic
gaze estimation network using  a very small amount of person-specific samples is very effective.

Notwithstanding the obtained improvements,
a limitation of our method is that the redirection synthesis is not good enough for large redirection angles.
It hinders further improvements of  gaze adaptation because generated samples can not cover the full space of gaze directions
and illumination conditions. 
We leave gaze redirection with larger angles and more illumination variabilities as  future work.

\mypartitle{Acknowledgement.}
This work was partly funded by the UBIMPRESSED project of the Sinergia interdisciplinary program of the Swiss National Science Foundation (SNSF), and
by the the European Unions Horizon 2020 research and innovation programme under grant agreement no. 688147 (MuMMER, mummer-project.eu).

{\small
\bibliographystyle{ieee}
\bibliography{egbib}
}

\end{document}